\documentclass[letterpaper, 10 pt, conference]{ieeeconf}
\IEEEoverridecommandlockouts                   
\usepackage{amssymb} 
\usepackage{amsmath} 
\usepackage{blkarray} 
\usepackage{wrapfig} 
\usepackage{array}
\usepackage{graphics} 
\usepackage{graphicx,subfigure}
\usepackage{epsfig}
\usepackage[ruled,linesnumbered,vlined]{algorithm2e}
\usepackage{multirow}
\usepackage{caption}
\usepackage{color,soul}
\usepackage{cite}
\usepackage{hyperref}
\usepackage[pscoord]{eso-pic}
\title{\LARGE \bf Robot Navigation in Crowds by Graph Convolutional Networks with Attention Learned from Human Gaze}

\author{Yuying Chen$^{*}$,  \  Congcong Liu$^{*}$, \  Ming Liu,  \  Bertram E. Shi  
\thanks{* These two authors contributed equally}
\thanks{All the authors are with the Hong Kong University of Science and Technology. {\tt\small \{ychenco, cliubh, eelium, eebert\}@ust.hk}}  %
}

\newcommand{\placetextbox}[3]{
  \AddToShipoutPictureFG*{
    \put(\LenToUnit{#1\paperwidth},\LenToUnit{#2\paperheight}){\vtop{{\null}\parbox{\textwidth}{#3}}}%
  }%
}%

\begin{document}
\placetextbox{0.09}{0.05}{\small{{\textcopyright}  2019 IEEE. Personal use of this material is permitted. Permission from IEEE must be obtained for all other uses, in any current or future media, including reprinting/republishing this material for advertising or promotional purposes, creating new collective works, for resale or redistribution to servers or lists, or reuse of any copyrighted component of this work in other works.}}

\maketitle
\thispagestyle{empty}
\pagestyle{empty}
\begin{abstract}

Safe and efficient crowd navigation for mobile robot is a crucial yet challenging task.
Previous work has shown the power of deep reinforcement learning frameworks to train efficient policies. However, their performance deteriorates when the crowd size grows. We suggest that this can be addressed by enabling the network to identify and pay attention to the humans in the crowd that are most critical to navigation.
We propose a novel network utilizing a graph representation to learn the policy.
We first train a graph convolutional network based on human gaze data that accurately predicts human attention to different agents in the crowd. Then we incorporate the learned attention into a graph-based reinforcement learning architecture. 
The proposed attention mechanism enables the assignment of meaningful weightings to the neighbors of the robot, and has the additional benefit of interpretability.
Experiments on real-world dense pedestrian datasets with various crowd sizes demonstrate that our model outperforms state-of-art methods by 18.4\% in task accomplishment and by 16.4\% in time efficiency.

\end{abstract}



\section{Introduction}

With the rapid development of artificial intelligence technologies, mobile robot navigation has many vital applications in crowded pedestrian environments such as hospitals, shopping malls, and canteens.
In these scenarios with dense crowds, navigating robots safely and efficiently is a crucial, yet still challenging, problem\cite{vemula2017modeling}.

Traditional approaches often treat pedestrians as simple dynamic obstacles and focus only on the next step \cite{van2008reciprocal,van2011reciprocal}. 
Since these approaches do not model human behavior, they result in behavior that can seem unnatural and short-sighted.
To achieve better long-term navigation, many research efforts have included reasoning about human intention and prediction of human trajectories before planing \cite{unhelkar2015human, kim2015brvo}.
However, doing prediction and planing separately may cause \textit{the freezing robot problem} when crowd density grows, because the planner believes every forward path will cause collision \cite{trautman2010unfreezing}. To address this problem, a key solution is to consider the impact of the motion of the robot on the crowds.

Current solutions can be divided into two categories: \textit{model based} and \textit{learning based}. Model-based methods mainly extend existing multi-agent collision avoidance solutions with explicit models of social interactions \cite{ferrer2013robot, mehta2016autonomous}. However, the model parameters need to be tuned for different application scenarios.
More recent research has used deep reinforcement learning (RL) successfully to learn efficient policies
that model the cooperation and interactions implicitly \cite{chen2018crowd, everett2018motion, chen2017decentralized}.

\begin{figure}
    \centering
    \includegraphics[width=0.9\columnwidth]{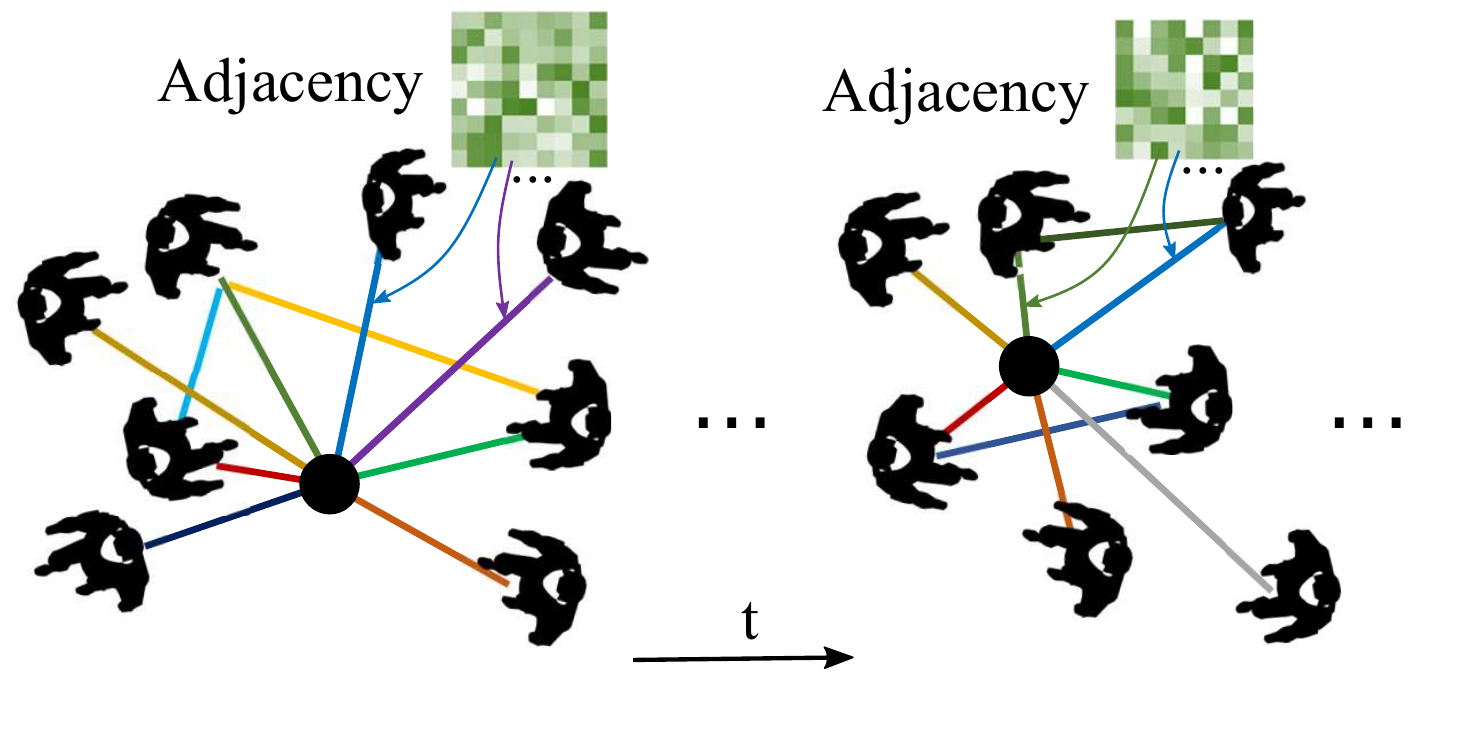}
  \caption{ We propose to represent the interactions among humans and robot as a graph with both nodes and edges varying over time. We use attention to modulate the adjacency matrix of a GCN to manage the feature processing and aggregation.\label{fig:idea}}
  \vspace{-1em}
\end{figure}

Deep RL methods proposed for crowd navigation have been model-based.  
They use reinforcement learning to learn a deep neural network that estimates the value function of a given robot-crowd configuration.
This value function is used in conjunction with a state transition function to perform action selection. 
To handle varying crowd sizes, 
the value estimation is often done by first estimating a fixed dimensional vector representing the robot-crowd state that combines information across the humans in the crowd, and then estimating the value from this fixed dimensional representation. 
Previously proposed approaches differ primarily in the structure of the networks used to encode the robot-crowd state and to estimate the corresponding value. 
There are several limitations of current models, 
which cause their performance to degrade when the crowd density increases \cite{chen2018crowd, everett2018motion}. 
First, existing models have considered only pairwise interactions between the robot and each human in the crowd.
By estimating these pairwise interactions independently,
these approaches do not completely capture the global and dynamic nature of crowd interactions.
Second, information about the crowd is obtained by combining information from these pairwise interactions by pooling \cite{chen2018crowd, gupta2018social, sadeghian2019sophie}, a maximum operation \cite{chen2017decentralized} or an LSTM that combines information sequentially from humans in the crowd according to their closeness to the robot \cite{everett2018motion}.
This does not completely capture important structural information about the geometric configuration of the crowd as a whole and the robot's relationship to it.

This work addresses these previous shortcomings in two ways.  First, we use a graph structure to represent the crowd state. Second, we use gaze data from humans performing a navigation task to learn a network that assigns different weights to different agents in the crowd according to their importance according to a measure of attention.

As depicted in Fig. \ref{fig:idea}, for robot navigation in dynamic crowds, it is natural to use a graph structure, which captures the relations (edges) between agents (nodes) to represent the crowd state.
Recently, Graph Convolutional Networks (GCNs) have been successfully applied with arbitrarily structured graphs in various areas, such as social networks \cite{chen2018fastgcn} and citation networks \cite{kipf2016semi}.
A GCN takes a feature matrix that represents attributes of each node as the input, and efficiently aggregates features from a neighborhood defined by an adjacency matrix.
One advantage of using a GCN to encode the state is that the interactions between nodes can be modulated easily by changing the adjacency matrix.

Previous research has shown that many visuomotor tasks such as driving \cite{liu2019gaze, chen2019gaze} and video games\cite{zhang2018agil} can benefit from the guidance of human attention as measured by gaze. Therefore, it is promising to investigate whether incorporating information about human attention information can also benefit mobile robot navigation.

We propose a novel attention-based RL approach to robot navigation using graph representation. We first train a two-layer GCN to predict human attention to the surrounding pedestrians while navigating in a simulator. The learned attention weights are incorporated into the adjacency matrix of a second GCN that is used to estimate the robot-crowd state. Finally, a deep neural net is used to estimate the corresponding value function.

There are three primary contributions of our work. First, we train an attention network that accurately predicts human attention in crowd navigation scenarios. Second, we propose a novel graph-represented reinforcement learning method for the crowd navigation task and demonstrate the architecture has significant benefits in learning an effective policy, which handles varying numbers of neighbors naturally and exhibits high extensibility. The architecture can be conveniently extended to more complex scenarios by adjusting the adjacency matrix. Third, we show that incorporating learned attention weights into 
a graph representation based reinforcement learning algorithm outperforms the state-of-the-art methods on real-world pedestrian trajectories data.

\section{Related Work}
\label{sec:related work}

In this section, we first summarize prior work in robot navigation. Then we introduce related work for graph-based deep learning. Finally, we review related work about the role of attention in different visuomotor learning tasks.

\subsection{Robot navigation in crowds}
Previous researchers have proposed many methods to solve the navigation problem. The Social Force Model \cite{helbing1995social} is one of the representative methods that has been successfully applied and extended in different environments \cite{ferrer2013robot, ferrer2017robot, chik2017gaussian}. Reciprocal velocity obstacles (RVO), which consider communications with other agents, were proposed in multi-agent navigation scenarios \cite{van2008reciprocal}. More recently, ORCA was proposed in \cite{van2011reciprocal}. It enables multiple robots to avoid collisions when navigating in a cluttered workspace. The main limitations of these model-based methods are that they require tedious parameter selection and that they may lead to unnatural robot behaviors, since they do not fully capture real human behavior.

Alternatively, imitation learning aims to learn optimal policies from human demonstrations directly. In the context of robot navigation, previous work has used imitation learning to obtain policies from raw 2D laser data \cite{long2017deep} or depth inputs \cite{tai2018socially} in a supervised way. Inverse reinforcement learning has also been applied to model human cooperative navigation behavior through a maximum entropy method \cite{pfeiffer2016predicting, kretzschmar2016socially}.

Deep reinforcement learning algorithms learn policies while the robot interacts with the environment through trial-and-error. For robot navigation, recent work has used reinforcement learning methods to learn policies from raw sensor inputs \cite{long2018towards} or agent-level representations of the environment \cite{everett2018motion, chen2018crowd}. Learning from raw sensor representation has the benefit that static and dynamic obstacles can be considered together through a single neural network. 
However, an agent-level representation can provide a richer high-level representation of pedestrian intent that is difficult to extract from raw sensor information. 
One challenge is the varying crowd size. Everett \textit{et al.} \cite{everett2018motion} converted the state of a variable-sized crowd to a fixed-length vector using an LSTM module that processed each pedestrian's state in descending order of their distance from the robot. However, the assignment of the importance according to distance is not always reasonable. 
For example, a pedestrian closely following a robot may be less important than a farther pedestrian in front of it.
A recent work \cite{chen2018crowd} adopted a self-attention module to assign different relative importances to different parts of the crowd.
In our work, we infer the relative importance of neighboring humans to the robot by learning attention weights from gaze data collected from humans performing a crowd navigation task.

\subsection{Graph representation learning}

Graph Convolutional Networks (GCNs) have attracted much attention for graph representation learning since they generalize the convolution operation to graph-structured data efficiently. GCNs have achieved remarkable successes in various research areas such as citation networks \cite{kipf2016semi}, social networks \cite{chen2018fastgcn} and material property prediction \cite{xie2018crystal}.
In the training of a GCN, a binary adjacency matrix is commonly used. However, the adjacency matrix can be weighted for adaptive and dynamic aggregation of neighbors' information.

A Graph Attention Network (GAT) \cite{velivckovic2017graph} is a variant of a GCN, which adopts a self-attention mechanism allowing the assignment of different importances to different nodes. Similarly, the Edge Attention based multi-relational GCN \cite{shang2018edge} replaces the adjacency matrix with a real-valued attention matrix, and learns the attention weights to optimize chemical property prediction.
In this work, we modulate the interactions in the crowds by using attention weights learned from human gaze data to determine the adjacency matrix.

\subsection{Human attention in visuomotor learning}

Human attention has been proved to be beneficial in learning effective policies for many visuomotor tasks, such as Atari games\cite{zhang2018agil} and autonomous driving \cite{liu2019gaze, chen2019gaze}.

Liu \textit{et al.} \cite{liu2019gaze} trained a conditional GAN to predict human attention and incorporated it into an end-to-end autonomous driving network. They added the predicted gaze map as an extra input to the network. Similarly, Zhang \textit{et al.} \cite{zhang2018agil} proposed an attention guided imitation learning network, which also treats the gaze map as additional image input in the task of eight Atari Games.

Chen \textit{et al.} proposed a novel form of dropout modulated by human gaze maps in imitation networks \cite{chen2019gaze}. The proposed gaze-modulated dropout leads to significant improvements in driving performance.

Here, we explore a novel way to incorporate human attention into deep reinforcement learning for the robot navigation task. We demonstrate significant benefits of our method in navigation performance.

\section{Methodology}
\label{sec:method}
As shown in Fig.\ref{fig:framework}, we designed a network to estimate the value function. The network utilized graph convolution, where the adjacency matrix is determined by another two-layer GCN which was learned based on human attention data. 
The value function is learned by reinforcement learning.

\subsection{Graph-based crowd representations}
Similar to the relations inside social networks, we represented the interactions among the robot and humans with graph structures. Each node denotes a human or robot. Edges denote the connections between them.

Handling the varying numbers and large densities of humans has always been a problem for crowd robot navigation. However, with the graph representations, the problem can be solved neatly. Unlike previous works that required an additional aggregation step to filter the features from humans or to combine them sequentially, we took advantage of the graph convolution layers. The output features of robot node aggregate information from the human nodes naturally.

A key feature of the graph convolutional layer is to generate a state embedding that contains information from related nodes. The layer-wise update rule is given by:
\begin{equation}
\mathbf{H}^{l+1} = \sigma \left ( \mathbf{A} \mathbf{H}^l \mathbf{W}^l \right )
\end{equation}
$\mathbf{H}^l$ contains the input node features of all nodes in the $l^{th}$ layer. It has size $N\times I$, where $N$ is the number of nodes in the graph and $I$ is the input feature length. $\mathbf{H}^{l+1} \in \mathbb{R}^{N\times O}$ contains the $O$ output features of each node.  $\mathbf{W}^l \in \mathbb{R}^{I\times O}$ is a weight matrix. The adjacency matrix $\mathbf{A} \in \mathbb{R}^{N \times N}$ defines the weights for combining the features from different nodes in the graph. We normalize it so that the sum of each row is equal to one. $\sigma(\cdot)$ denotes the activation function, which we choose to be the Relu.

\begin{figure}[t!]
\centering
{
\includegraphics[width=0.65\columnwidth]{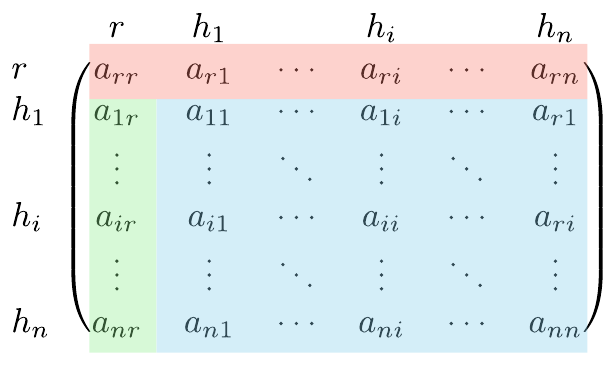}
\caption{Adjacency matrix of the graph network for robot-crowd interactions. The red area considers the robot attention on the humans as well as itself. The green area considers the visibility of the robot to the human. The blue area considers the interaction among the crowd. The subscripts 'r' and 'h' indicate the robot and human respectively. \label{fig:adj}}
}
\vspace{-1em}
\end{figure}

The adjacency matrix reflects the topology of the graph structure. Zero values represent no connections between corresponding nodes. Higher values represent stronger connections. For the graph of interactions among the robot and humans, the meaning of the adjacency matrix is shown in Fig.\ref{fig:adj}. The adjacency matrix is divided into three functional areas. The red area and green area are related to the human-robot interactions. Elements in the red area represent the importance that the robot assigns to the humans and itself, which can also be interpreted as attention. Elements in the green area indicate the influence of the robot on the humans. It may vary from human to human according to their mental state or concentration. The worst case is the humans not noticing the robot, which leads to zero values. The blue area considers the interactions inside the crowd (human-human interactions). The adjacency matrix can be designed to reflect different situations conveniently and efficiently. 

In this work, we focused on the red area and investigated a method to set the weights assigned to the robot and the humans. For the green and blue areas, we set the values as shown in Equation \ref{eqn:att_assign1} and Equation \ref{eqn:att_assign2} respectively. 
\begin{eqnarray}
  a_{ir} = 1/2, i=1,2,...n  \label{eqn:att_assign1}\\
  a_{ij} = \left\{\begin{matrix}
            0, i\neq j
            \\
            1/2, i=j
            \end{matrix}\right.  \label{eqn:att_assign2}
\end{eqnarray}

We propose to utilize learned human attention to modulate the adjacency values in the red part. The proposed method is referred to as Gaze modulated GCN based RL (\textit{G-GCNRL}). As baselines, we trained models with distance-related weights (\textit{D-GCNRL}) and uniform weights (\textit{U-GCNRL}). 
For the \textit{D-GCNRL}, the adjacency values were set using the attention weights in Equation \ref{eqn:blue}, described later. For the \textit{U-GCNRL}, the adjacency values in the red part were all equal and summed to one.

\subsection{Graph-based deep V-learning}
\begin{figure}
    \centering
    \includegraphics[width=0.6\columnwidth]{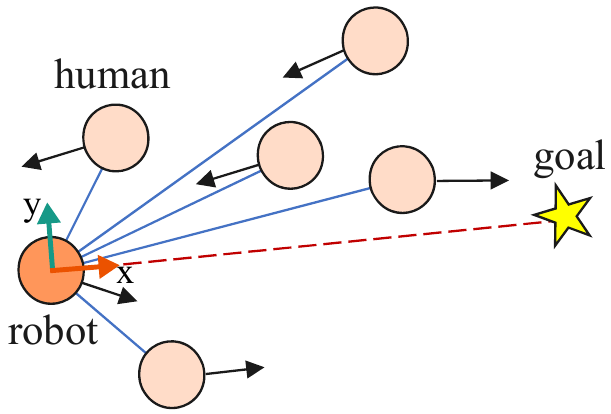}
  \caption{Topology of graph for crowd feature learning. The black arrows indicate the moving directions of robot and humans. The red and green arrows indicate the $x$ and $y$ axes of the robot-centric coordinates. The blue lines represent the graph edges.\label{fig:topo}}
  \vspace{-0.5em}
\end{figure}

\begin{figure}
    \includegraphics[width=0.9\columnwidth]{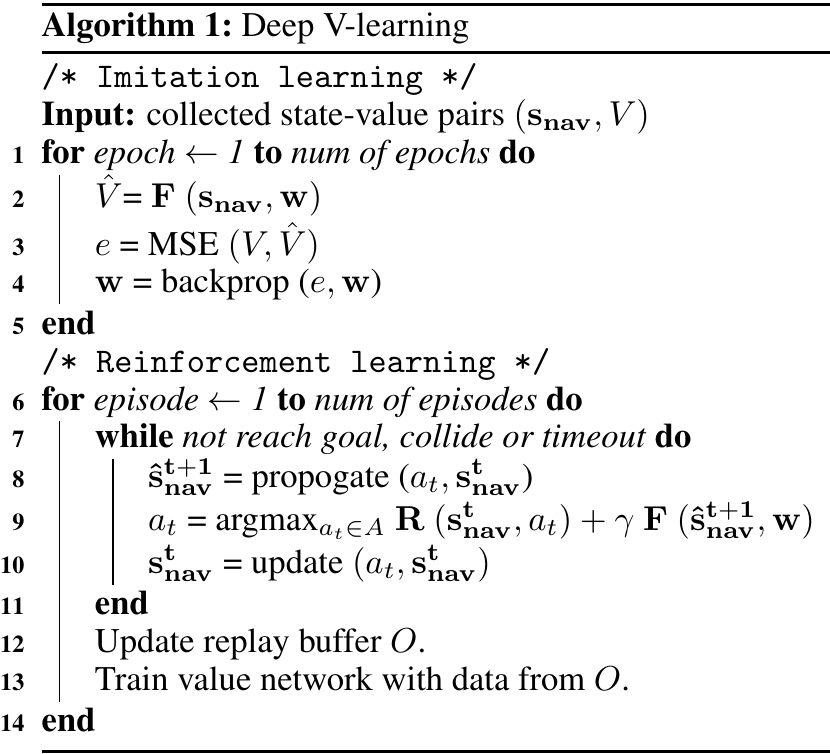} 
     \vspace{-3em}
\end{figure}

As mentioned above, the graph representation is beneficial for aggregating information about the crowd. We utilized the GCN to extract a representation of the influence of the crowd state on the robot. Interactions between the robot and humans were considered. The graph topology is shown in Fig.\ref{fig:topo}. The input to each node ($s_{nav}$) was designed to incorporate both robot and node information:

\begin{eqnarray}
    &s_{nav} = [s_{r}, s_{n}], \\
  &s_{r} = [d_g, v_{pref}, \theta, r_r, v_{r_x}, v_{r_y}], \\
  &s_{n} = [x, y, v_{n_x}, v_{n_y}, r_n, d_r, r_n + r_r].
\end{eqnarray}
$s_{r}$ represents the state of the robot, including distance from the goal, preferred velocity, moving direction, radius of footprint and velocities. $s_{n}$ represents the state of the node (robot $n=0$, or human $n=\{1,...,N\}$), including positions, velocities, radius, distance to the robot and sum of the radii. All the velocities and positions are in a robot-centric coordinate system where the $x$ axis points in the direction of the goal.

\begin{figure*}
     \centering
    \includegraphics[width=1.7\columnwidth]{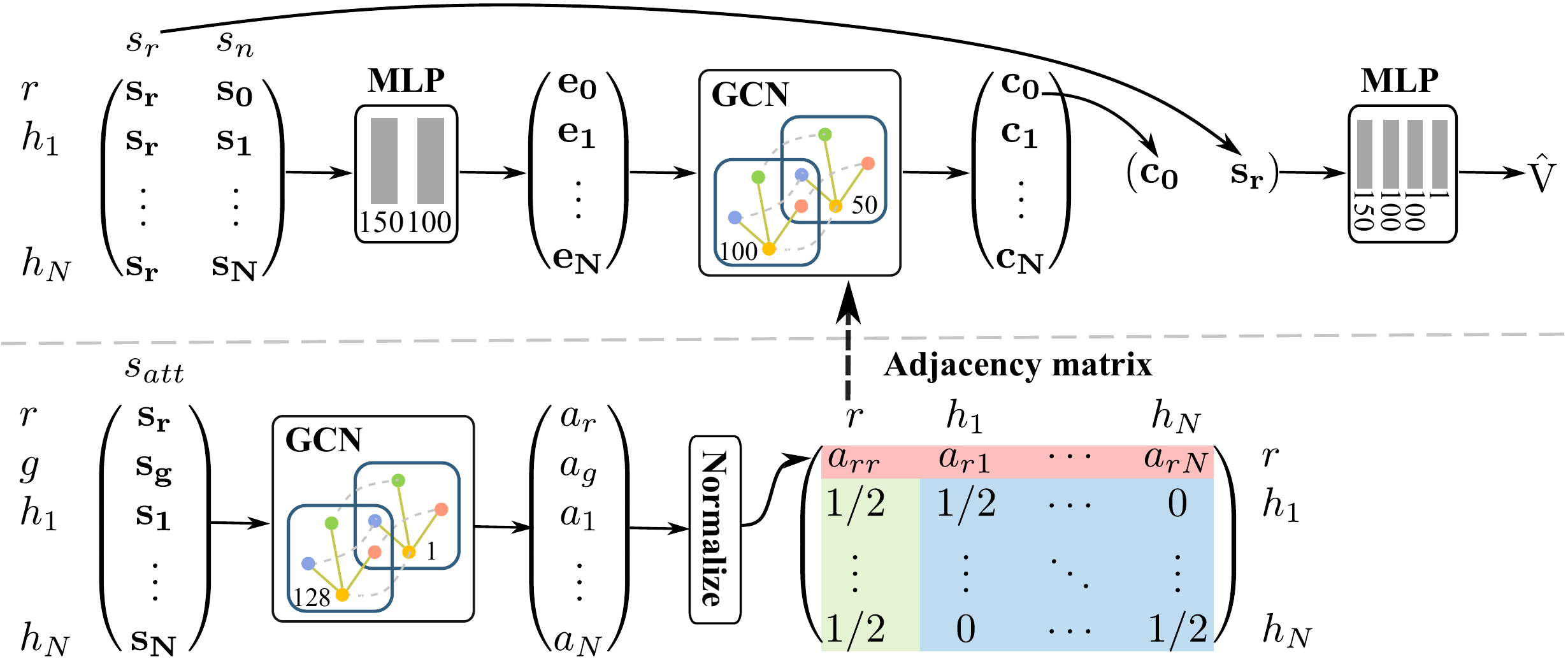}
  \caption{Framework of graph based V-learning with learned human attention. The upper part and the lower part show the graph based V-learning and the attention network respectively. They are connected through the adjacency matrix of the GCN contained in the policy network. Rows of matrix marked with $r$, $g$ and $h$ indicate robot, goal, and human relevant content, respectively. Columns of matrix marked with $s_{\{r,n,att\}}$ have the contents following their definitions accordingly. The extracted crowd features (marked as $\mathbf{c_0}$) are from the output of robot node. The value prediction is marked as $\hat{V}$.     \label{fig:framework}}
  \vspace{-1em}
 \end{figure*}

As shown in the top of Fig.\ref{fig:framework}, the input state of each node first passes through a multilayer perceptron (MLP) to increase the feature length to 100 for sufficient expressive power. The output features $\mathbf{\left \{ e_i \right \}_{i=0}^N}$ are sent into the GCN, which has two graph convolutional layers. We extracted the output features $C_0$ of robot node as the crowd features. The robot node aggregates information about the humans through the graph convolutions. After that, the extracted crowd features together with the robot status are concatenated and sent into the value network for V-learning.

One of the key set of parameters in the GCN is the adjacency matrix, as this captures the influence of the states of different nodes on their neighbors. Similar to the graph attention network (GAT) \cite{velivckovic2017graph}, we set the values of the adjacency matrix based on values of attention, which were computed by another neural network. However, unlike the GAT, we used a different network architecture, as described in the next section, and trained the attention weights to mimic these estimated from human gaze data, rather than to capture the concept of self-attention.

As shown in Algorithm 1, the value network learned the value of a robot state, which guides the action selection of the robot. We followed the problem formulation and reward function design in \cite{chen2018crowd}\cite{chen2017decentralized}.  Given space limitations, some details are omitted. 
The learning process of the value network consists of two stages. At the first stage, the network was trained with supervision. It learned to imitate the demonstrations of ORCA, which drives the robot through the crowd to the destination. For data collection, both successful cases and failure cases were recorded. We collected data in the simulation environment from the augmented crowd dataset \textit{Students001} \cite{lerner2007crowds}. It contains real human trajectories with high crowd density. We simulated the sequential state update of humans as the robot navigated inside the crowd. 

After initialization by imitation learning, the value network was trained further by reinforcement learning. We used the same environment for training. For each episode, we sampled multiple actions from the action space ($A$) at every step. We selected the action with the largest value. States of the robot and humans were then updated, and the state-value pairs stored in the replay buffer. 
We kept a size limit on the buffer so the old pairs were replaced by new ones gradually. For imitation learning, we collected 3000 episodes. For reinforcement learning, we ran 20000 episodes and set the buffer size to 100000 state-value pairs.  

\subsection{GCN to predict human attention}

We used a GCN to compute the attention weights. Similar to the crowd feature extraction, we established graphs having a star topology with the robot as the central connection point. Elements likely to affect the attention distribution of robot directly were considered. Therefore, humans, the robot and the goal were all included as the graph nodes. The connections between the robot and humans as well as between the robot and its goal were the graph edges. 
We used a graph representation to extract crowd features, as it is a meaningful and efficient way for data aggregation. The problem of learning attention is a node regression problem, since we must assign an attention weight to every node.

Input features of the robot, humans, and robot goal were represented as:
\begin{equation}
  s_{att} = [x, y, v_x, v_y],
\end{equation}
where $[x, y]$ and $[v_x, v_y]$ denote the position and velocity in the robot-centric coordinate system.

Each node generated an attention weight as its output. The network contained two GCN layers with output sizes of 128 and 1. The forward model of our attention network follows \cite{kipf2016semi}: 
 
\begin{equation}
\mathbf{H^2} = \mathrm{softmax} \left(\mathbf{A} \mathrm{ReLU}\left ( \mathbf{A}\mathbf{H}^0\mathbf{W}^0 \right ) \mathbf{W}^1 \right )
\end{equation}

Here, $\mathbf{W}^0$ and $\mathbf{W}^1$ denote the input-to-hidden and hidden-to-output weight matrices respectively. $\mathbf{H}^0$ and $\mathbf{H}^2$ denote the input and output of the attention network respectively. The adjacency matrix \textbf{A} had a similar form as in \textit{U-GCNRL}. Elements in the first row were all equal. 

We used ReLU as the activation function. Dropout was applied after each graph convolutional layer with a drop probability of 0.5. 
The network output went through a softmax function to ensure that the attention weights summed to one. 

The attention network $\mathbf{G}$ was learned with supervision. The ground truth was generated from human gaze data. Gaze data collected while subjects attempted to navigate through the crowd to the goal using a joystick, while viewing the scene from overhead. We reproduced the scene in the real world crowd dataset \cite{lerner2007crowds}. 
For training, we used the state-attention weight pairs collected in \textit{Students001} environment. We collected gaze data for multiple starts with varying crowd sizes and densities, resulting in around 1700 state and attention pairs. To generate the ground truth attention weights, we first created a Gaussian mixture density over space by placing a Gaussian with standard deviation equals to two degree of visual angle at each gaze point in a temporal window from -0.1 to +0.1 seconds around the current time point. We assigned attention weights to each node by sampling from the density according to the node location, and normalized so that the attention weights summed to one.
We used the L1 loss function and trained the network for 400 epochs until convergence.

\section{Experimental Results}
\label{sec:result}
In this section, we evaluated our attention network and our \textit{G-GCNRL} model. Due to the dynamic nature of the experiment, we encourage readers to view the video\footnote{Available from the supplementary files or the website: \url{https://sites.google.com/view/gazenav}} for a better understanding. 

\subsection{Human attention modelling}

\begin{figure}
    \centering
    {
    \includegraphics[width=0.9\columnwidth]{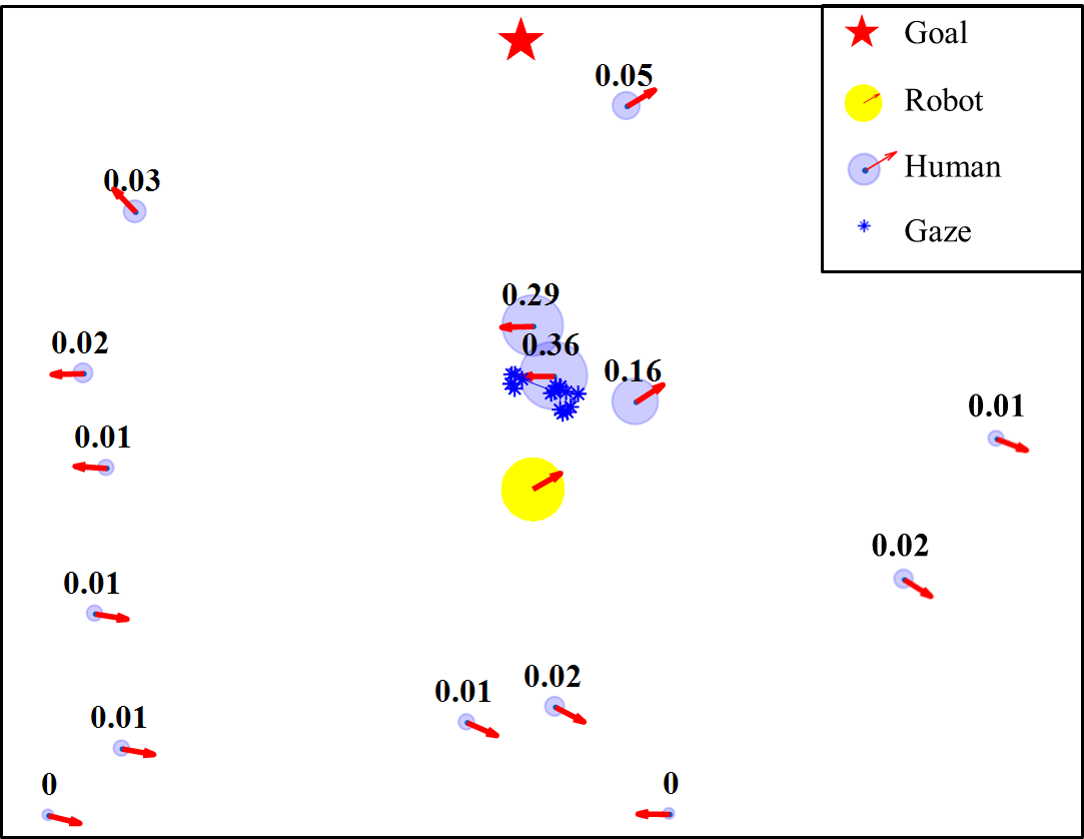}
  \caption{Learned attention weights from our attention network are shown by the radius of the purple circle surrounding each human. Measured gaze trajectory is shown in blue. Red arrows indicate the instantaneous velocity of each agent. \label{fig:attention_weights}}}
  \vspace{-1em}
\end{figure} 

\begin{figure*}
    \centering
    \includegraphics[width=0.7\textwidth]{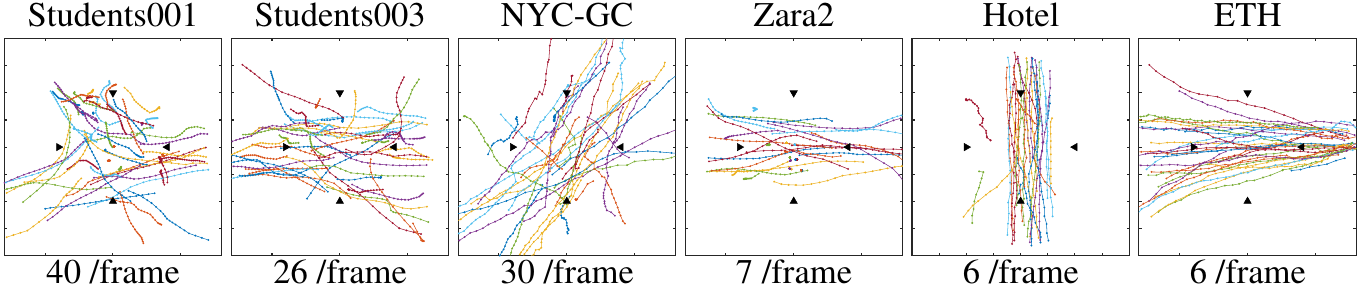}
    \caption{Dataset for simulation environment setting up. The colorful lines show the trajectories of pedestrians. For better visualization, the trajectories shown here have gone through ten times down sampling. The black triangles show the starting positions and goals of the robot, which are 4m away from the center. The average number of humans per frame are shown below each image. Students001, 003 and Zara2 are from \cite{lerner2007crowds}. NYC-GC is from \cite{zhou2012understanding}. Hotel and ETH are from \cite{pellegrini2009you}.}
    \label{fig:scene}
    \vspace{-1em}
\end{figure*}

Figure \ref{fig:attention_weights} compares the estimated attention weights and the ground truth gaze trajectory for a test crowd scene. 
The robot's path to the goal was blocked by two humans moving to the left. 
To avoid them, the human steered the robot so that it followed the human moving to the right.
The learned attention weights are largest for the humans near the gaze of the human operator. These correspond to the humans being avoided and the human being followed.
Thus, the learned attention network can necessarily infer which human in the crowd are most important.

To evaluate the predicted attention weights quantitatively, we calculated the similarity between predicted attention weights and the ground truth weights
using two standard metrics: the Kullback-Leibler divergence (KL) and the Correlation Coefficient (CC). Smaller KL and larger CC denote better similarity.
For comparison, we also consider two alternate ways to assign attention weights: based on distance and based on \textit{SARL} \cite{chen2018crowd}. Intuitively, humans close to the robot should exert a stronger influence on the robot. Therefore, we considered assigning attention weights that decayed with the distance between the human and the robot: as shown in Equation \ref{eqn:blue}.
\begin{equation}
      a_{ij} = \frac{e^{-d_{ij}^2/\sigma ^2}}{\sum_{j=1}^{n} e^{-d_{ij}^2/\sigma ^2}}, i=1,2,..n \label{eqn:blue}
\end{equation}
$\sigma^2$ was set to 2, which was found empirically by brute force search to maximize success rate of the \textit{D-GCNRL} after imitation learning.
The other baseline we chose was the weights obtained from the self-attention module of \textit{SARL}.

\label{tab:gaze_pred}
\begin{table}
\centering
\begin{tabular}{lcc}
\hline
                            & KL-D & CC   \\ \hline
Predicted attention weights & 0.49 & 0.74 \\ \hline
Distance related weights    & 0.99 & 0.61 \\ \hline
Self-attention weights in \cite{chen2018crowd}   & 0.92 & 0.63 \\ \hline
\end{tabular}
\caption{Similarity of attention estimates. KL-D denotes Kullback-Leibler divergence and CC denotes Correlation Coefficient.}
\vspace{-1em}
\end{table}

The attention weights predicted by our network closely match the ground truth weights. As shown in Table. \ref{tab:gaze_pred}, the KL divergence for the predicted attention weights is 50.5\% smaller than for the distance related weights and 46.7\% smaller than for the self-attention weights. Similarly, the CC for the predicted attention weights is 21.3\% larger than for the distance related weights and 17.5 \% larger than for the self-attention weights.

\subsection{Graph represented reinforcement learning with attention}
\begin{figure*}
    \centering
    \includegraphics[width=0.95\textwidth]{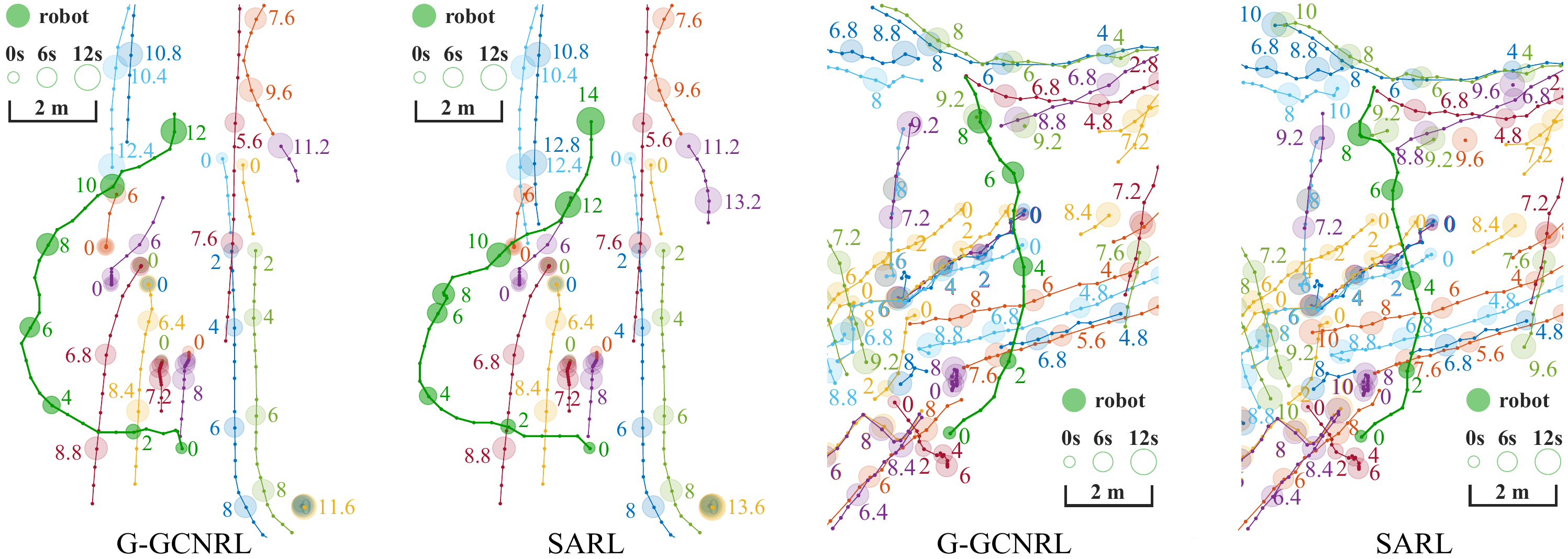}
    \caption{Trajectory of robot running in the simulated environments, with both simple case (Zara2) and hard case (NYC-GC) showed here. The highlighted green lines showed the trajectory of robot and the other colors showed the humans'. The timestamp of each circle are marked. }
    \label{fig:nav_traj}
\end{figure*}

\begin{table*}[]
\centering
\begin{tabular}{lcccccccccccc}
\hline
        & \multicolumn{6}{c}{Success Rate}                  & \multicolumn{6}{c}{Navigation Time}               \\ \hline
        & Student003 & NYC-GC & Zara2 & Hotel & ETH  & \textbf{AVG}  & Student003 & NYC-GC & Zara2 & Hotel & ETH  & \textbf{AVG}  \\ \hline
SARL    & 0.692       & 0.358  & 0.815  & 0.581  & 0.657 & 0.621 & 13.1       & 13.8   & 12.9  & 13.4  & 13.9 & 13.4 \\ \hline
SA-GCNRL & 0.616       & 0.431   & 0.838  & 0.683  & 0.782 & 0.670 & 12.0       & 12.0   & 13.2  & 12.8  & 13.0 & 12.6 \\ \hline
G-GCNRL  & \textbf{0.753}       & \textbf{0.453}   & \textbf{0.936}  & \textbf{0.703}  & \textbf{0.831} & \textbf{0.735} & \textbf{11.2}       & \textbf{11.8}   & \textbf{10.9}  & \textbf{11.1}  & \textbf{11.1} & \textbf{11.2} \\ \hline
\end{tabular}
\caption{Comparison to the state-of-the-art, \textit{SARL}. }
\label{tab:cp2sarl}
\end{table*}

\begin{table*}[]
\centering
\begin{tabular}{lcccccccccccc}
\hline
         & \multicolumn{6}{c}{Success Rate}                  & \multicolumn{6}{c}{Navigation Time}               \\ \hline
         & Student003 & NYC-GC & Zara2 & Hotel & ETH  & \textbf{AVG}  & Student003 & NYC-GC & Zara2 & Hotel & ETH  & \textbf{AVG}  \\ \hline
G-GCNRL  & \textbf{0.753}       & \textbf{0.453}   & \textbf{0.936}  & \textbf{0.703}  & \textbf{0.831} & \textbf{0.735} & \textbf{11.2}       & \textbf{11.8}   & 10.9  & 11.1  & 11.1 & 11.2 \\ \hline
SA-GCNRL & 0.616       & 0.431   & 0.838  & 0.683  & 0.782 & 0.670 & 12.0       & 12.0   & 13.2  & 12.8  & 13.0 & 12.6 \\ \hline
D-GCNRL  & 0.556       & 0.405   & 0.876  & 0.699  & 0.790 & 0.665 & 12.7       & 13.8   & 11.5  & 9.3   & 10.3 & 11.5 \\ \hline
U-GCNRL  & 0.671       & 0.387   & 0.928  & 0.687  & 0.827 & 0.700 & 11.2       & 12.0   & \textbf{10.1}  & \textbf{9.3 }  & \textbf{10.2} & \textbf{10.6} \\ \hline
\end{tabular}
\caption{Additional ablation study to show the advantage of the attention weights trained based on human gaze data.}
\label{tab:ab2attention}
\end{table*}

\begin{table*}[]
\centering
\begin{tabular}{lcccccccccccc}
\hline
         & \multicolumn{6}{c}{Success Rate}                  & \multicolumn{6}{c}{Navigation Time}               \\ \hline
         & Student003 & NYC-GC & Zara2 & Hotel & ETH  & AVG  & Student003 & NYC-GC & Zara2 & Hotel & ETH  & AVG  \\ \hline
SA-GCNRL & 0.616       & \textbf{0.431}   & \textbf{0.838}  & \textbf{0.683}  & \textbf{0.782} & \textbf{0.670} & \textbf{12.0}       & \textbf{12.0}   & 13.2  & \textbf{12.8}  & \textbf{13.0} & \textbf{12.6} \\ \hline
SARL     & \textbf{0.692}       & 0.358   & 0.815  & 0.581  & 0.657 & 0.621 & 13.1       & 13.8   & \textbf{12.9} & 13.4  & 13.9 & 13.4 \\ \hline
U-GCNRL  & \textbf{0.671}       & \textbf{0.387}   & \textbf{0.928}  & \textbf{0.687}  & \textbf{0.827} & \textbf{0.700} & \textbf{11.2}       & \textbf{12.0}   & \textbf{10.1}  & \textbf{9.3}   & \textbf{10.2} & \textbf{10.6} \\ \hline
UARL    & 0.591       & 0.310   & 0.816  & 0.477  & 0.636 & 0.566 & 12.7       & 13.1   & 14.0  & 14.0  & 15.0 & 13.8 \\ \hline
\end{tabular}
\caption{Additional ablation study to show the advantage of the graph structure.}
\label{tab:ab2gcn}
\end{table*}
To verify the performance of the proposed method and the advantage of our two innovations (the introduction of the graph structure and the use of attention weights learned from human gaze), we first compared our \textit{G-GCNRL} to the state-of-the-art \textit{SARL} (Table \ref{tab:cp2sarl}). Then we performed additional ablation studies to validate the performance gains of each of the two innovations individually and in combination (Tables \ref{tab:ab2attention} - \ref{tab:ab2gcn}). 
All tests were conducted in the environments shown in Fig. \ref{fig:scene} with random selection of the starting frames. We set the robot to start at each black triangle and go to the opposite one. We collected the results of each model over 2500 trials. 


\subsubsection{Comparing with SARL}

Similar to our \textit{G-GCNRL} approach, \textit{SARL} assigns different importances to different humans. 
There are two differences: we use a GCN to estimate the value function and use human gaze data to learn the attention weights.
Therefore, we implemented another model (\textit{SA-GCNRL}) to demonstrate the advantages of using the GCN alone.

\textit{SA-GCNRL}: We replaced the learned human attention in \textit{G-GCNRL} with the self-attention of \textit{SARL}. 

As shown in Table \ref{tab:cp2sarl}, \textit{SA-GCNRL} achieves on average a 7.9\% higher success rate and a 6.0\% shorter navigation time than \textit{SARL}. 

The \textit{G-GCNRL} to \textit{SA-GCNRL} comparison shows the advantage of using the learned attention weights learned from human gaze data. As shown in Table \ref{tab:cp2sarl}, \textit{G-GCNRL} achieves a 9.7\% higher success rate and an 11.1\% shorter navigation time than \textit{SA-GCNRL}.

Taken together, the results demonstrate the superiority of both graph-based feature aggregation and of the learned human attention. Both combined lead to significant improvement, 18.4\% higher success rate and 16.4\% shorter navigation time.

Fig. \ref{fig:nav_traj} shows two examples of trajectories generated by \textit{G-GCNRL} and \textit{SARL} in both a simple environment (Zara2) and a dense environment (NYC-GC). 
In the simple environment, we can see that the robot (green circle) controlled by \textit{SARL} almost stopped between 6 and 8 seconds, whereas the velocity of the robot controlled by \textit{G-GCNRL} was much more constant. It reached the final goal in 12.4 s rather than the 14 s taken by \textit{SARL}.
For the dense environment, the path taken by the \textit{G-GCNRL} robot after 6 s is much straighter than the path taken by \textit{SARL} robot. The \textit{G-GCNRL} robot achieves its goal in 9.2 s, rather than the 10 s taken by the \textit{SARL} robot.
For both environments, \textit{G-GCNRL} achieves the goal faster than \textit{SARL}. 
It can be observed that \textit{SARL} took more conservative policies. The robot waited somewhere or changed direction unnecessarily. \textit{G-GCNRL} drove the robot more smoothly. 

\subsubsection{Ablation study for learned human attention}

To evaluate the benefits of the learned attention, we compared the results of \textit{G-GCNRL} with two other methods (\textit{D-GCNRL} and \textit{U-GCNRL}). \textit{D-GCNRL} weights the importance of humans by the distance from the robot. 
\textit{U-GCNRL} set equal attention weights for all nodes. 

Using attention weights learned from human gaze improved the success rate significantly. As shown in Table \ref{tab:ab2attention}, the success rate of \textit{G-GCNRL} was 10.5\% higher than that of \textit{D-GCNRL} and 5.0\% higher than that of \textit{U-GCNRL}.  
However, if the goal was achieved, the navigation time of \textit{G-GCNRL} was not shorter. \textit{U-GCNRL} generally achieved the shortest navigation time.

\subsubsection{Ablation study for graph structure}
To evaluate the benefits using the GCN structure to estimate the crowd state, we ran two comparisons. Both confirm the advantage of the graph structure.

First, we compared \textit{U-GCNRL}, a GCN where all attention weights were equal with \textit{UARL}, a \textit{SARL}-like network where all of the attention weights were equal. As shown in Table \ref{tab:ab2gcn}, \textit{U-GCNRL} achieves 23.7\% higher success rate and 23.2\% shorter navigation time.

Second, we compared \textit{SA-GCNRL}, a GCN with attention weights determined by \textit{SARL}, and \textit{SARL}. As shown in Table \ref{tab:ab2gcn}, \textit{SA-GCNRL} achieves 7.9\% higher success rate and 6.0\% shorter navigation time. 

\subsubsection{Performance in different environments}
Of all the testing environments, NYC-GC is the most complex, with on average 30 humans per frame. In addition, the humans were moving in multiple directions. All models obtained the worst performance in NYC-GC (Table \ref{tab:cp2sarl}-\ref{tab:ab2gcn}).
The models with uniform attention (\textit{UARL} and \textit{U-GCNRL}) performed much worse than corresponding models with self attention (\textit{SARL}) or learned human attention (\textit{G-GCNRL}). For crowds with low densities, the performance of the methods was more similar. This is expected, since when the crowd size is small, it is relatively easy to keep track of all the humans. However, when the crowd size is large, this quickly becomes overwhelming. Attention becomes critical in enabling the system to focus on the most important parts of the crowd. In all environments, the \textit{G-GCNRL} achieves highest success rate. 

\section{Conclusion}
\label{sec:conclusion}

In this paper, we presented a crowd navigation method for mobile robots and demonstrated its efficacy on real-world dense pedestrian data. The proposed method outperforms the state-of-the-art.
There are two key innovations in our work. The first innovation is the introduction of a graph convolutional network (GCN) for reinforcement learning to integrate information about the environmental context of the robot. 
The GCN makes our approach natively adaptable to varying crowd sizes in a principled way.
The influence of different agents can be controlled by changing the adjacency matrix. 
The second innovation is the introduction of an attention network trained using human gaze data for assigning adjacency values. 
The two innovations enhance the performance of the network may also be useful in other applications. 

\bibliographystyle{IEEEtran}
\bibliography{example}

\begin{thebibliography}{10}
\providecommand{\url}[1]{#1}
\csname url@rmstyle\endcsname
\providecommand{\newblock}{\relax}
\providecommand{\bibinfo}[2]{#2}
\providecommand\BIBentrySTDinterwordspacing{\spaceskip=0pt\relax}
\providecommand\BIBentryALTinterwordstretchfactor{4}
\providecommand\BIBentryALTinterwordspacing{\spaceskip=\fontdimen2\font plus
\BIBentryALTinterwordstretchfactor\fontdimen3\font minus
  \fontdimen4\font\relax}
\providecommand\BIBforeignlanguage[2]{{%
\expandafter\ifx\csname l@#1\endcsname\relax
\typeout{** WARNING: IEEEtran.bst: No hyphenation pattern has been}%
\typeout{** loaded for the language `#1'. Using the pattern for}%
\typeout{** the default language instead.}%
\else
\language=\csname l@#1\endcsname
\fi
#2}}

\bibitem{vemula2017modeling}
A.~Vemula, K.~Muelling, and J.~Oh, ``Modeling cooperative navigation in dense
  human crowds,'' in \emph{2017 IEEE International Conference on Robotics and
  Automation (ICRA)}.\hskip 1em plus 0.5em minus 0.4em\relax IEEE, 2017, pp.
  1685--1692.

\bibitem{van2008reciprocal}
J.~Van~den Berg, M.~Lin, and D.~Manocha, ``Reciprocal velocity obstacles for
  real-time multi-agent navigation,'' in \emph{2008 IEEE International
  Conference on Robotics and Automation}.\hskip 1em plus 0.5em minus
  0.4em\relax IEEE, 2008, pp. 1928--1935.

\bibitem{van2011reciprocal}
J.~Van Den~Berg, S.~J. Guy, M.~Lin, and D.~Manocha, ``Reciprocal n-body
  collision avoidance,'' in \emph{Robotics Research}.\hskip 1em plus 0.5em
  minus 0.4em\relax Springer, 2011, pp. 3--19.

\bibitem{unhelkar2015human}
V.~V. Unhelkar, C.~P{\'e}rez-D'Arpino, L.~Stirling, and J.~A. Shah,
  ``Human-robot co-navigation using anticipatory indicators of human walking
  motion,'' in \emph{2015 IEEE International Conference on Robotics and
  Automation (ICRA)}.\hskip 1em plus 0.5em minus 0.4em\relax IEEE, 2015, pp.
  6183--6190.

\bibitem{kim2015brvo}
S.~Kim, S.~J. Guy, W.~Liu, D.~Wilkie, R.~W. Lau, M.~C. Lin, and D.~Manocha,
  ``Brvo: Predicting pedestrian trajectories using velocity-space reasoning,''
  \emph{The International Journal of Robotics Research}, vol.~34, no.~2, pp.
  201--217, 2015.

\bibitem{trautman2010unfreezing}
P.~Trautman and A.~Krause, ``Unfreezing the robot: Navigation in dense,
  interacting crowds,'' in \emph{2010 IEEE/RSJ International Conference on
  Intelligent Robots and Systems}.\hskip 1em plus 0.5em minus 0.4em\relax IEEE,
  2010, pp. 797--803.

\bibitem{ferrer2013robot}
G.~Ferrer, A.~Garrell, and A.~Sanfeliu, ``Robot companion: A social-force based
  approach with human awareness-navigation in crowded environments,'' in
  \emph{2013 IEEE/RSJ International Conference on Intelligent Robots and
  Systems}.\hskip 1em plus 0.5em minus 0.4em\relax IEEE, 2013, pp. 1688--1694.

\bibitem{mehta2016autonomous}
D.~Mehta, G.~Ferrer, and E.~Olson, ``Autonomous navigation in dynamic social
  environments using multi-policy decision making,'' in \emph{2016 IEEE/RSJ
  International Conference on Intelligent Robots and Systems (IROS)}.\hskip 1em
  plus 0.5em minus 0.4em\relax IEEE, 2016, pp. 1190--1197.

\bibitem{chen2018crowd}
C.~Chen, Y.~Liu, S.~Kreiss, and A.~Alahi, ``Crowd-robot interaction:
  Crowd-aware robot navigation with attention-based deep reinforcement
  learning,'' \emph{arXiv preprint arXiv:1809.08835}, 2018.

\bibitem{everett2018motion}
M.~Everett, Y.~F. Chen, and J.~P. How, ``Motion planning among dynamic,
  decision-making agents with deep reinforcement learning,'' in \emph{2018
  IEEE/RSJ International Conference on Intelligent Robots and Systems
  (IROS)}.\hskip 1em plus 0.5em minus 0.4em\relax IEEE, 2018, pp. 3052--3059.

\bibitem{chen2017decentralized}
Y.~F. Chen, M.~Liu, M.~Everett, and J.~P. How, ``Decentralized
  non-communicating multiagent collision avoidance with deep reinforcement
  learning,'' in \emph{2017 IEEE International Conference on Robotics and
  Automation (ICRA)}.\hskip 1em plus 0.5em minus 0.4em\relax IEEE, 2017, pp.
  285--292.

\bibitem{gupta2018social}
A.~Gupta, J.~Johnson, L.~Fei-Fei, S.~Savarese, and A.~Alahi, ``Social gan:
  Socially acceptable trajectories with generative adversarial networks,'' in
  \emph{Proceedings of the IEEE Conference on Computer Vision and Pattern
  Recognition}, 2018, pp. 2255--2264.

\bibitem{sadeghian2019sophie}
A.~Sadeghian, V.~Kosaraju, A.~Sadeghian, N.~Hirose, H.~Rezatofighi, and
  S.~Savarese, ``Sophie: An attentive gan for predicting paths compliant to
  social and physical constraints,'' in \emph{Proceedings of the IEEE
  Conference on Computer Vision and Pattern Recognition}, 2019, pp. 1349--1358.

\bibitem{chen2018fastgcn}
J.~Chen, T.~Ma, and C.~Xiao, ``Fastgcn: fast learning with graph convolutional
  networks via importance sampling,'' \emph{arXiv preprint arXiv:1801.10247},
  2018.

\bibitem{kipf2016semi}
T.~N. Kipf and M.~Welling, ``Semi-supervised classification with graph
  convolutional networks,'' \emph{arXiv preprint arXiv:1609.02907}, 2016.

\bibitem{liu2019gaze}
C.~Liu, Y.~Chen, L.~Tai, H.~Ye, M.~Liu, and B.~E. Shi, ``A gaze model improves
  autonomous driving,'' in \emph{Proceedings of the 11th ACM Symposium on Eye
  Tracking Research \& Applications}.\hskip 1em plus 0.5em minus 0.4em\relax
  ACM, 2019, p.~33.

\bibitem{chen2019gaze}
Y.~Chen, C.~Liu, L.~Tai, M.~Liu, and B.~E. Shi, ``Gaze training by modulated
  dropout improves imitation learning,'' \emph{arXiv preprint
  arXiv:1904.08377}, 2019.

\bibitem{zhang2018agil}
R.~Zhang, Z.~Liu, L.~Zhang, J.~A. Whritner, K.~S. Muller, M.~M. Hayhoe, and
  D.~H. Ballard, ``Agil: Learning attention from human for visuomotor tasks,''
  in \emph{Proceedings of the European Conference on Computer Vision (ECCV)},
  2018, pp. 663--679.

\bibitem{helbing1995social}
D.~Helbing and P.~Molnar, ``Social force model for pedestrian dynamics,''
  \emph{Physical Review E}, vol.~51, no.~5, p. 4282, 1995.

\bibitem{ferrer2017robot}
G.~Ferrer, A.~G. Zulueta, F.~H. Cotarelo, and A.~Sanfeliu, ``Robot social-aware
  navigation framework to accompany people walking side-by-side,''
  \emph{Autonomous Robots}, vol.~41, no.~4, pp. 775--793, 2017.

\bibitem{chik2017gaussian}
S.~F. Chik, C.~F. Yeong, E.~L.~M. Su, T.~Y. Lim, F.~Duan, J.~T.~C. Tan, P.~H.
  Tan, and P.~J.~H. Chin, ``Gaussian pedestrian proxemics model with social
  force for service robot navigation in dynamic environment,'' in \emph{Asian
  Simulation Conference}.\hskip 1em plus 0.5em minus 0.4em\relax Springer,
  2017, pp. 61--73.

\bibitem{long2017deep}
P.~Long, W.~Liu, and J.~Pan, ``Deep-learned collision avoidance policy for
  distributed multiagent navigation,'' \emph{IEEE Robotics and Automation
  Letters}, vol.~2, no.~2, pp. 656--663, 2017.

\bibitem{tai2018socially}
L.~Tai, J.~Zhang, M.~Liu, and W.~Burgard, ``Socially compliant navigation
  through raw depth inputs with generative adversarial imitation learning,'' in
  \emph{2018 IEEE International Conference on Robotics and Automation
  (ICRA)}.\hskip 1em plus 0.5em minus 0.4em\relax IEEE, 2018, pp. 1111--1117.

\bibitem{pfeiffer2016predicting}
M.~Pfeiffer, U.~Schwesinger, H.~Sommer, E.~Galceran, and R.~Siegwart,
  ``Predicting actions to act predictably: Cooperative partial motion planning
  with maximum entropy models,'' in \emph{2016 IEEE/RSJ International
  Conference on Intelligent Robots and Systems (IROS)}.\hskip 1em plus 0.5em
  minus 0.4em\relax IEEE, 2016, pp. 2096--2101.

\bibitem{kretzschmar2016socially}
H.~Kretzschmar, M.~Spies, C.~Sprunk, and W.~Burgard, ``Socially compliant
  mobile robot navigation via inverse reinforcement learning,'' \emph{The
  International Journal of Robotics Research}, vol.~35, no.~11, pp. 1289--1307,
  2016.

\bibitem{long2018towards}
P.~Long, T.~Fanl, X.~Liao, W.~Liu, H.~Zhang, and J.~Pan, ``Towards optimally
  decentralized multi-robot collision avoidance via deep reinforcement
  learning,'' in \emph{2018 IEEE International Conference on Robotics and
  Automation (ICRA)}.\hskip 1em plus 0.5em minus 0.4em\relax IEEE, 2018, pp.
  6252--6259.

\bibitem{xie2018crystal}
T.~Xie and J.~C. Grossman, ``Crystal graph convolutional neural networks for an
  accurate and interpretable prediction of material properties,''
  \emph{Physical Review Letters}, vol. 120, no.~14, p. 145301, 2018.

\bibitem{velivckovic2017graph}
P.~Veli{\v{c}}kovi{\'c}, G.~Cucurull, A.~Casanova, A.~Romero, P.~Lio, and
  Y.~Bengio, ``Graph attention networks,'' \emph{arXiv preprint
  arXiv:1710.10903}, 2017.

\bibitem{shang2018edge}
C.~Shang, Q.~Liu, K.-S. Chen, J.~Sun, J.~Lu, J.~Yi, and J.~Bi, ``Edge
  attention-based multi-relational graph convolutional networks,'' \emph{arXiv
  preprint arXiv:1802.04944}, 2018.

\bibitem{lerner2007crowds}
A.~Lerner, Y.~Chrysanthou, and D.~Lischinski, ``Crowds by example,'' in
  \emph{Computer Graphics Forum}, vol.~26, no.~3.\hskip 1em plus 0.5em minus
  0.4em\relax Wiley Online Library, 2007, pp. 655--664.

\bibitem{zhou2012understanding}
B.~Zhou, X.~Wang, and X.~Tang, ``Understanding collective crowd behaviors:
  Learning a mixture model of dynamic pedestrian-agents,'' in \emph{2012 IEEE
  Conference on Computer Vision and Pattern Recognition}.\hskip 1em plus 0.5em
  minus 0.4em\relax IEEE, 2012, pp. 2871--2878.

\bibitem{pellegrini2009you}
S.~Pellegrini, A.~Ess, K.~Schindler, and L.~Van~Gool, ``You'll never walk
  alone: Modeling social behavior for multi-target tracking,'' in \emph{2009
  IEEE 12th International Conference on Computer Vision}.\hskip 1em plus 0.5em
  minus 0.4em\relax IEEE, 2009, pp. 261--268.

\end{thebibliography}
\end{document}